\def\year{2022}\relax
\documentclass[letterpaper]{article} 
\usepackage{aaai22}  
\usepackage{times}  
\usepackage{helvet}  
\usepackage{courier}  
\usepackage[hyphens]{url}  
\usepackage{graphicx} 
\usepackage{natbib}  
\usepackage{caption} 
\DeclareCaptionStyle{ruled}{labelfont=normalfont,labelsep=colon,strut=off} 
\usepackage{algorithm}
\usepackage[noend]{algorithmic}
\usepackage{amssymb}
\usepackage{amsthm}
\usepackage{amsmath}
\usepackage{xspace}
\usepackage{color}
\usepackage{etoolbox,siunitx}
\robustify\bfseries
\usepackage{booktabs}
\usepackage{multicol}

\usepackage{newfloat}
\usepackage{listings}

\newcommand{\videoset}{\mathcal{S}}
\newcommand{\ansset}{\mathcal{A}}
\newcommand{\V}{{S}}
\newcommand{\vidclip}{{s}}
\newcommand{\Q}{{Q}}
\newcommand{\A}{{A}}
\newcommand{\Ap}{{A_{\text{pred}}}}
\newcommand{\Agt}{{A_{\text{gt}}}}
\newcommand{\f}{f}
\newcommand{\fo}{f^o}
\newcommand{\fa}{f^a}
\newcommand{\depth}{D}
\newcommand{\graph}{\mathcal{G}}
\newcommand{\gtf}{\graph_{3.5D}}
\newcommand{\class}{\mathcal{C}}
\newcommand{\feat}{F}
\newcommand{\ftd}{F_{3.5D}}
\newcommand{\fhd}{F^H_{3.5D}}
\newcommand{\fqd}{F^Q_{3.5D}}
\newcommand{\query}{\mathbf{Q}}
\newcommand{\key}{\mathbf{K}}
\newcommand{\val}{\mathbf{V}}

\DeclareMathOperator*{\softmax}{softmax}
\DeclareMathOperator*{\concat}{||}

\DeclareMathOperator*{\MLP}{MLP}

\newcommand{\name}{(2.5+1)D\xspace}
\newcommand{\nameTxr}{(2.5+1)D-Transformer\xspace}

\newcommand{\vset}{\mathcal{V}}
\newcommand{\nodes}{V}
\newcommand{\node}{v}

\newcommand{\eset}{\mathcal{E}}

\newcommand{\cc}[1]{\textcolor{red}{#1}}

\newcommand{\kernel}{\kappa}
\newcommand{\Kernel}{\mathbf{K}}

\newcommand{\lnorm}[1]{\left\|{#1}\right\|_1}

\newcommand{\reals}[1]{\mathbb{R}^{#1}}

\newcommand{\enorm}[1]{\left\|{#1}\right\|}

\DeclareMathOperator{\argmin}{arg\,min}
\newcommand{\set}[1]{\left\{#1\right\}}

\DeclareMathOperator{\bbox}{bbox}

\DeclareMathOperator{\criteria}{C}

\DeclareMathOperator{\roipool}{ROIPool}
\DeclareMathOperator{\itd}{I3D}
\DeclareMathOperator{\match}{match}
\DeclareMathOperator{\ancestor}{ancestor}

\floatstyle{ruled}
\newfloat{listing}{tb}{lst}{}
\floatname{listing}{Listing}
\title{(2.5+1)D Spatio-Temporal Scene Graphs for Video Question Answering}
\author{
Anoop Cherian\qquad Chiori Hori\qquad Tim K. Marks\qquad Jonathan Le Roux
}
\affiliations{
    Mitsubishi Electric Research Labs (MERL), Cambridge, MA\\
   \{cherian, chori, tmarks, leroux\}@merl.com
}

\begin{document}

\maketitle
\begin{abstract}
Spatio-temporal scene-graph approaches to video-based reasoning tasks, such as video question-answering (QA), typically construct such graphs for every video frame. These approaches often ignore the fact that videos are essentially sequences of 2D ``views'' of events happening in a 3D space, and that the semantics of the 3D scene can thus be carried over from frame to frame. Leveraging this insight, we propose a \name scene graph representation to better capture the spatio-temporal information flows inside the videos. Specifically, we first create a 2.5D (pseudo-3D) scene graph by transforming every 2D frame to have an inferred 3D structure using an off-the-shelf 2D-to-3D transformation module, following which we register the video frames into a shared \name spatio-temporal space and ground each 2D scene graph within it. Such a \name graph is then segregated into a static sub-graph and a dynamic sub-graph, corresponding to whether the objects within them usually move in the world. The nodes in the dynamic graph are enriched with motion features capturing their interactions with other graph nodes. Next, for the video QA task, we present a novel transformer-based reasoning pipeline that embeds the \name graph into a spatio-temporal hierarchical latent space, where the sub-graphs and their interactions are captured at varied granularity. To demonstrate the effectiveness of our approach, we present experiments on the NExT-QA and AVSD-QA datasets. Our results show that our proposed \name representation leads to faster training and inference, while our hierarchical model showcases superior performance on the video QA task versus the state of the art. 
\end{abstract}

\section{Introduction}
Recent advances in deep learning have made it possible to think beyond individual domains, such as computer vision and natural language processing, and consider tasks that are at their intersections. Visual question answering (VQA) is one such task that has witnessed a significant attention lately~\cite{antol2015vqa,anderson2018bottom,wu2017visual,jang2017tgif,geng2021dynamic,chen2020counterfactual,ghosh2019generating}. While earlier approaches to this task used holistic visuo-textual representations~\cite{dang2021hierarchical,antol2015vqa}, it was found that decomposing a visual scene into its constituents (and their relationships) provided a better reasoning pipeline~\cite{anderson2018bottom,johnson2015image,krishna2017visual,dornadula2019visual}, perhaps because of the possibility for an easier disentanglement of the scene objects relevant to the given question. Such a disentanglement naturally leads to a graph representation of the scene, usually called a scene graph~\cite{johnson2015image}. Using such a graph representation allows one to use the powerful machinery of graph neural networks for VQA, and has demonstrated significant promise~\cite{li2019know,li2019relation,pan2020spatio,geng2021dynamic}.

While visual scene graphs were originally proposed for image-based tasks, there have been direct adaptations of this data structure for video-based reasoning problems~\cite{geng2021dynamic,chatterjee2021visual,pan2020spatio,herzig2019spatio}. Usually, in such problems, scene graphs are constructed for every video frame, followed by an inter-frame representation learning to produce holistic video level features for reasoning. However, having scene graphs for every frame may be redundant and could even become computationally detrimental for longer video sequences. Taking a step back, we note that videos are essentially 2D views of a 3D space in which various events happen temporally, and representing the scene in a 4D spatio-temporal space could thus potentially avoid such representational redundancies. Furthermore, object properties such as permanence~\cite{shamsian2020learning} could be handled more effectively in a 3D space, as each object (that is visible in some video frame) gets a location therein, thereby disentangling the camera views from its spatial location~\cite{tung20203d,girdhar2019cater}. Using such a 3D representation thus would provide a natural way to avoid occlusions, which is a significant problem when working with 2D scene graphs.

Motivated by the above insight, we explore a novel spatio-temporal scene graph representation, where the graph nodes are not grounded on individual video frames, instead are mapped to a shared 3D world coordinate frame. While there are approaches in computer vision that could produce such a common 3D world~\cite{multiview}, such methods usually assume: i) that the scene is static, without dynamic objects,  ii) that the camera calibration information is known, or iii) that multiple overlapping views of the same scene are available; none of which may exist for arbitrary (internet) videos typically used in VQA tasks. Fortunately, there have been several recent advancements in 3D reconstruction from 2D images, such as~\cite{ranftl2019towards,fu2018deep}; these works take as input an image and produces a realistic pseudo-3D structure for the image scene, typically called a 2.5D image. For every video frame, we leverage such a 2.5D reconstruction to impart an approximate 3D location for each graph node, thereby producing a \name spatio-temporal scene graph.

\begin{figure*}[ht]
    \centering
    \includegraphics[width=16cm,trim={0.4cm 6.3cm 0cm 3cm},clip]{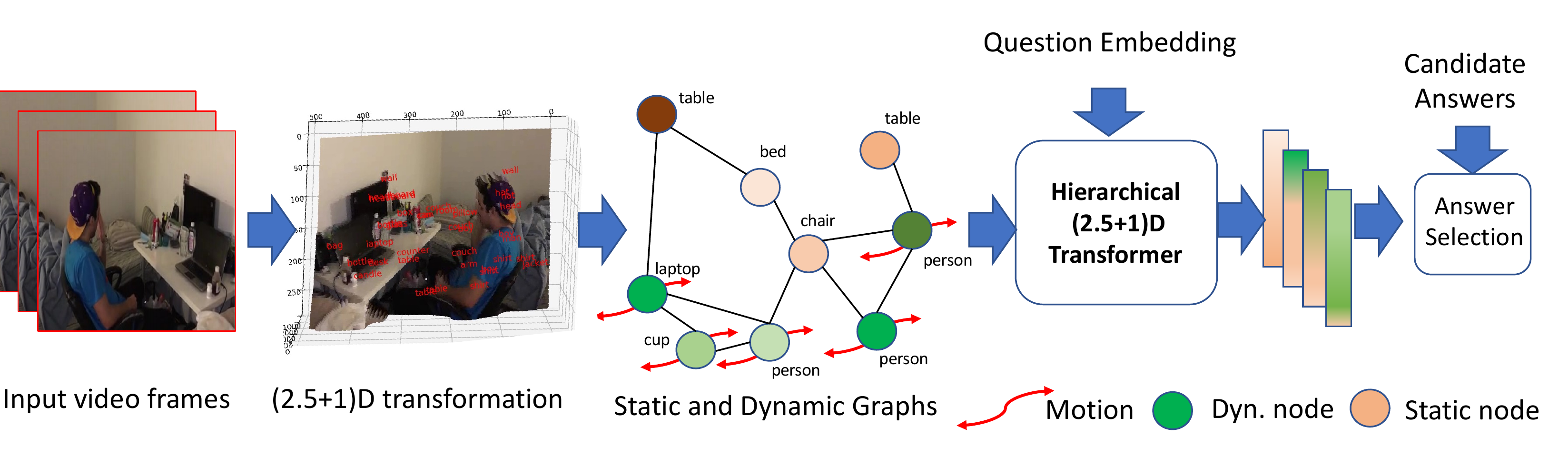}
    \caption{A schematic illustration of our proposed \name video QA reasoning pipeline.}
    \label{fig:pipeline}
\end{figure*}
A technical challenge with the above \name scene graph representation is that each graph is still specific to a video frame, and is not registered to a shared space. Such a registration is confounded by the fact that objects in the scene may move from frame to frame. To this end, we propose to: (i) split the \name scene graph into a static 2.5D sub-graph and a dynamic \name sub-graph, depending on whether the class of the underlying scene graph node usually moves in scenes (e.g., a \emph{person} class is dynamic, while a \emph{table} class is considered static), (ii) merge the graph nodes corresponding to the static sub-graph based on their 3D spatial proximity across frames, thereby removing the node redundancy, and (iii) retain the nodes of the dynamic sub-graph from the original scene graph. As the dynamic sub-graph nodes are expected to not only capture the frame-level semantics, but to also potentially involve object actions (e.g., person \emph{picking} a bottle), we enrich each dynamic graph node with motion features alongside its object-level feature representation. Thus, our proposed \name scene graph representation approximately summarizes the spatio-temporal activity happening in a scene in a computationally efficient framework.

The use of such a \name graph representation allows for developing rich inference schemes for VQA tasks. For example, to capture the interaction of a person with a static object in the scene, the inference algorithm needs to attend to regions in the \name graph where the spatio-temporal proximity between the respective graph nodes is minimized. Leveraging this intuition, we propose a hierarchical latent embedding of the \name graph where the graph edges are constructed via varied spatio-temporal proximities, thereby capturing the latent embeddings of the graph at multiple levels of granularity. We use such a graph within a Transformer reasoning pipeline~\cite{vaswani2017attention} and conditioned on the VQA questions to retrieve the correct answer. 

To validate the effectiveness of our approach, we present experiments on two recent video QA datasets, namely: (i) the NExT-QA dataset~\cite{xiao2021next} and (ii) the QA task of the audio-visual scene aware dialog (AVSD) dataset~\cite{alamri2019audio}. Our results on these datasets show that our proposed framework leads to about 4x speed up in training, while pruning 25--50\% graph nodes, and showcases superior QA accuracy against recent state-of-the-art methods.  

\section{Related Work}
 We note that visual question answering has been a very active research area in the recent times, and thus interested readers may refer to excellent surveys, such as~\cite{teney2018tips,wu2017visual}. In the following, we restrict our literature review to prior methods that are most similar to our contributions.
 
\noindent\textbf{Scene graphs for QA:} Since the seminal work of~\cite{johnson2015image} in using scene graphs as a rich representation of an image scene, there have been extensions of this idea for video QA and captioning tasks~\cite{herzig2019spatio,wang2018non,jang2017tgif,tsai2019video,girdhar2019video}. Spatio-temporal scene graphs are combined with a knowledge distillation objective for video captioning in~\cite{pan2020spatio}. Similarly, video scene graphs are combined with multimodal Transformers for video dialogs and QA in~\cite{geng2021dynamic}. In~\cite{Jiang_Gao_Guo_Zhang_Xiang_Pan_2019}, a graph alignment framework is proposed that uses graph co-attention between visual and language cues for better video QA reasoning. In~\cite{fan2019heterogeneous}, a multi-step reasoning pipeline is presented that attends to visual and textual memories. We note that scene graphs have been explored for various action recognition tasks as well. For example, video action graphs are presented in~\cite{bar2020compositional,rashid2020action,Wang2018videos}. Action Genome~\cite{ji2019action,cong2021spatial} characterizes manually annotated spatio-temporal scene graphs for action recognition. In contrast to these prior methods, we seek a holistic and potentially minimal representation of a video scene via pseudo-3D scene graphs for the QA task.

\noindent\textbf{3D scene graphs:} Very similar to our motivations towards a comprehensive scene representation, 3D scene graphs have been proposed in~\cite{armeni20193d}. However, their focus is on efficient annotation and collection of such graphs from 3D sensors. Similarly, more recent efforts such as~\cite{zhang2021exploiting,wu2021scenegraphfusion} are also targeted at improvising the efficiency of constructing a 3D scene graph from RGBD scans, while our focus is on constructing pseudo-3D graphs leveraging recent advancements in 2D-to-3D methods. We note that while precise 3D scene graphs may be important for several tasks such as robot navigation or manipulation, they need not be required for reasoning tasks such as what we consider in this paper, and for such tasks approximate 3D reasoning may be sufficient. We also note that 3D graphs have been explored for video prediction tasks in~\cite{tung20203d}, however in a very controlled setting.

\noindent\textbf{Graph Transformers:} Similar to our contribution, connections between graphs and Transformers have been explored previously. For example,~\cite{choromanski2021graph} has explored long-range attention using kernelized Transformers, while~\cite{tsai2019Transformer} presents a kernel view of Transformer attention, and Bello presents a long range attention using lambda layers that captures both position and content interactions~\cite{bello2021lambdanetworks}. While there are some similarities between these works and ours in the use of kernels and positional details in computing the similarity matrix within a Transformer, our objective and goals are entirely different from these works. Specifically, our proposed architecture is to represent a pseudo-3D scene at multiple levels of spatio-temporal granularity for a reasoning task, which is entirely different from the focus of these prior works.
\section{Proposed Method}
In this section, we first present our setup for constructing \name scene graphs for a given video sequence, then explain our hierarchical spatio-temporal Transformer-based graph reasoning pipeline. See Fig.~\ref{fig:pipeline} for an overview of our framework.

\subsection{Problem Setup}
We assume that we have access to a set of $N$ training video sequences, $\videoset=\set{\V_1, \V_2, \dots, \V_N}$, where the $i$-th video consists of $n_i$ frames. In the following, we eliminate the subscripts for simplicity, and use $\V$ to denote a generic video sequence from $\videoset$ that has $n$ frames. We assume that each video $\V$ is associated with at least one question $\Q$, which is an ordered tuple of words from a predefined vocabulary (tokenized and embedded suitably). We define the task of video QA as that of retrieving a predicted answer, $\Ap$, from a collection of $\ell$ possible answers, $\ansset=\set{\A_1, \A_2, \dots, \A_\ell}$. Of these $\ell$ answers, we denote the ground-truth answer as $\Agt$. We propose to represent $\V$ as a \name spatio-temporal scene graph. The details of this process are described in the following subsections.

\subsection{2D Scene Graph Construction}
Let $\graph=(\vset, \eset)$ be a scene graph representation of a video sequence $\V$ of length $n$ frames, where $\vset = \nodes_1\cup \nodes_2\cup\cdots \cup\nodes_n$ denotes the set of nodes, each $\nodes_t$ denotes the subset of nodes associated with frame $t$, and $\eset\subseteq \vset\times \vset$ denotes the set of graph edges (which are computed as part of our hierarchical Transformer framework explained later). To construct the scene graph $\graph$, we follow the standard pipeline using an object detector. Specifically, we first extract frames from the video sequence and pass each frame as input to a Faster R-CNN (FRCNN) object detection model~\cite{ren2015faster}. The FRCNN implementation that we use is pre-trained on the Visual Genome dataset~\cite{anderson2018bottom} and can thus detect 1601 object classes, which include a broad array of daily-life indoor and outdoor objects. In every frame, the FRCNN model detects $m$ objects, each of which is represented by a graph node $\node$ that contains a tuple of FRCNN outputs $(\fo_\node, c_\node, \bbox_\node)$, where $\fo_\node$ is the object's neural representation, $c_\node$ is its label in the Visual Genome database, and $\bbox_\node$ denotes its bounding box coordinates relative to the respective frame. Thus, for a video sequence with $n$ frames, we will have $mn$ graph nodes.\footnote{While this may appear not too big a memory footprint, note that each visual feature $\fo_\node$ is usually a 2048D vector. Thus, with $m=36$, videos of length $n \approx 50$ frames, and a batch size of 64, we would need about 15GB of GPU memory for forward propagation alone.} However, as alluded to above, several of these graph nodes may be redundant, thus motivating us to propose our \name scene graphs.

\subsection{\name Scene Graphs}
Suppose $\depth\!:\reals{h\times w\times 3}\to\reals{h\times w\times 4}$ denotes a neural network model that takes as input an RGB image and produces as output an RGBD image, where the depth is estimated. For a video frame (image) $I$, further let $d_{I}:\reals{2}\to\reals{3}$ map a 2D pixel location $(x, y)$ to a respective 3D coordinate, denoted $p=(x,y,z)$. To implement $\depth$, we use an off-the-shelf pre-trained 2D-to-3D deep learning framework. While there are several options for this network~\cite{fu2018deep,li2020unsupervised}, we use the MiDAS model~\cite{ranftl2019towards}, due to its ease of use and its state-of-the-art performance in estimating realistic depth for a variety of real-world scenes. 
For a scene graph node $\node \in \nodes_t$ extracted from video frame $t$ (image $I_t$), let $\overline{\bbox}_\node$ denote the centroid of the node's detected bounding box. To enrich the scene graph with \name spatio-temporal information, we expand the representation of node $\node$ to include depth and time by updating the tuple for $\node$ to be $(\fo_\node, c_\node, \bbox_\node, p_\node, t)$, where $p_\node = d_{I_t}(\overline{\bbox}_\node)$ can be interpreted as the 3D centroid of the bounding box.
We denote the enriched graph as~$\graph_{3.5D}$.

\subsection{Static and Dynamic Sub-graphs}
While the nodes in $\gtf$ are equipped with depth, they are still grounded in every video frame, which is potentially wasteful. This is because many of these nodes may correspond to objects in the scene that seldom move in the real world. If we can identify such objects, then we can prune their redundant scene graph nodes. To this end, we segregated the Visual Genome classes into two distinct categories,  namely (i) a category $\class_s$ of static scene objects, such as \emph{table}, \emph{tree}, \emph{sofa}, \emph{television}, etc., and (ii) a category $\class_d$ of dynamic objects, such as \emph{people}, \emph{mobile}, \emph{football}, \emph{clouds}, etc. While the visual appearance of a static object may change from frame to frame, we assume that its semantics do not change and are sufficient for reasoning about the object in the QA task. However, such an assumption may not hold for a dynamic object, such as a \emph{person} who may interact with various objects in the scene, and each interaction needs to be retained for reasoning. Using this class segregation, we split the graph $\gtf$ into two distinct scene graphs, $\graph_s$ and $\graph_d$, corresponding to whether the object label $c_\node$ of a node $\node\in\vset$ belongs to $\class_s$ or $\class_d$, respectively. 

Our next subgoal is to register $\gtf$ in a shared 3D space. There are two key challenges in such a registration, namely that (i) the objects in the scene may move, and (ii) the camera may move. These two problems can be tackled easily if we extract registration features only from the static sub-graph nodes of the frames. Specifically, if there is camera motion, then one may find a frame-to-frame 3D projection matrix using point features, and then use this projection matrix to spatially map all the graph nodes (including the dynamic nodes) into a common coordinate frame. 

While this setup is rather straightforward, we note that the objects in the static nodes are only defined by their bounding boxes, which are usually imprecise. Thus, to merge two static nodes, we first consider whether the nodes are from frames that are sufficiently close in time, with the same object labels, and with the intersection over union (IoU) of their bounding boxes above a threshold $\gamma$. Two nodes  $\node_t,\node_{t'}\in\graph_s$, from frames with timestamps $t\neq t'$ (where $|t-t'|<\delta$) are candidates for merging if the following criterion $\criteria$ is met: 
\begin{equation}
   \criteria(\node_t, \node_{t'}) \!:=\! \left(c_{\node_t} \!=\! c_{\node_{t'}}\right) 
   \wedge\mathrm{IoU}(\bbox_{\node_t}, \bbox_{\node_{t'}})>\gamma. 
   \label{eq:criterion}
\end{equation}
If a static node $\node_t$ has multiple candidate nodes in the previous $\delta$ frames that satisfy criterion~\eqref{eq:criterion}, the candidate with the nearest 3D centroid is selected as the matching node that will be merged:
\begin{equation}
   \match(\node_t) = \!
   \underset{\begin{array}{c}{\node_{t'} \in \nodes^s_{t-\delta}\cup \cdots \cup\nodes^s_{t-1}} \\ 
   \text{such that }\criteria(\node_t, \node_{t'}) = 1 \end{array}}{\argmin}
   \! \enorm{p_{\node_t} - p_{\node_{t'}}}\! ,
   \label{eq:match}
\end{equation}
where $\nodes^s_t = \{\node_t \in \nodes_t \;|\; \node_t \in \graph_s\}$ denotes the set of all static nodes from frame $t$. Since \eqref{eq:match} chooses the best match from the past $\delta$ frames, rather than just from frame $t-1$, it can tolerate more noise in the estimates of the depth and the bounding boxes associated with the graph nodes. 

We can apply this matching process recursively in order to determine larger equivalence classes of matched nodes to be merged, where an equivalence class is defined as the set of all nodes that share a single common ancestor. We accomplish this by looping over the frames $t$ in temporal order, where for each node $\node_t$ for which $\match(\node_t)$ exists, we assign 
$\ancestor(\node_t) = \ancestor\big(\match(\node_t)\big)$. This procedure is detailed in Algorithm~\ref{alg:ancestors}. Finally, for each ancestor, all nodes that share that ancestor are merged into a single node. The feature $\fo_\node$ associated with the new node $\node$ is obtained by averaging the features from all of the nodes that were merged into it. We use the 3D coordinate $p$ of the parent node for all the child nodes that are merged into it. Let $\graph_{s'}$ denote the new reduced version of $\graph_s$ after each equivalence class of matched nodes has been merged into one node.

\begin{algorithm}[tb]
\caption{Identifying common ancestors for merging}
\label{alg:ancestors}
\begin{algorithmic}
\FOR{$\node_1 \in \nodes^s_1$}
\STATE $\ancestor(\node_1) := \node_1$
\ENDFOR
\FOR{$t = 2$ to $n$}
\FOR{$\node_t \in \nodes^s_t$}
\IF{$\match(\node_t)$ exists}
\STATE $\ancestor(\node_t) := \ancestor\bigl(\match(\node_t)\bigr)$
\ENDIF
\ENDFOR
\ENDFOR
\end{algorithmic}
\end{algorithm}
\subsection{Motion Features}
To recap, so far we have segregated the graph $\gtf$ into $\graph_s$ and $\graph_d$, where the nodes of $\graph_s$ have been pruned and registered into a common shared 3D space to form an updated graph $\graph_{s'}$, while the spatial locations of the nodes in the dynamic graph $\graph_d$ have been updated via transformation matrices produced from $\graph_{s'}$ into the same coordinate frame. An important step missing in our framework is that the dynamic sub-graph that we have constructed so far is still essentially a series of graph nodes produced from FRCNN, which is a static object detection model -- that is, the nodes are devoid of any \emph{action} features that are perhaps essential in capturing how a dynamic node \emph{acts} within itself and on its environment (defined by the static objects). To this end, we propose to incorporate motion features into the nodes of the dynamic graph. Specifically, we use the I3D action recognition neural network~\cite{carreira2017quo}, pre-trained on the Kinetics-400 dataset to produce convolutional features from short temporal video clips. These features are then ROI-pooled using the (original) bounding boxes associated with the dynamic graph nodes. Suppose $\fa_{\node_t}=\roipool(\itd(\vidclip_t), \bbox_{\node_t})$, where $\vidclip_t$ denotes a short video clip around the $t$-th video frame of a video $\V$, then we augment the FRCNN feature vector by concatenating the object and action features as $\f^{oa}_\node \leftarrow \fo_\node \concat \fa_\node$, for all $\node\in\nodes_d$, where $||$ denotes feature concatenation.

\subsection{Hierarchical Graph Embedding}
Using our \name scene graph thus constructed, we are now ready to present our video QA reasoning setup. As the questions in a QA task may need reasoning at various levels of abstraction (e.g., Q:~\emph{what is the color of a person's shirt?} A: \emph{red}, Q:~\emph{why did the boy cry?}: A:~\emph{Because the ball hit him}, etc.), we decided to design our reasoning pipeline so that it can capture such a hierarchy. To set the stage, let us review a few basics on Transformers in our context. In the sequel, we assume the set of nodes in $\gtf$ is given by: $\nodes'=\nodes_{s'}\cup \nodes_d$.

\noindent\textbf{Transformers:} Suppose $\feat\in \reals{r\times |\nodes'|}$ denotes a matrix of features computed from the static and dynamic graph nodes of a video $\V$ via projecting their original features into latent spaces of dimensionality $r$ using multi-layer perceptrons, $\MLP_s$ and $\MLP_d$; i.e., $\feat=\MLP_s(\fo_{\nodes_{s'}}) \concat \MLP_d(\f^{oa}_{\nodes_d})$. If $\query^i_F, \key^i_F, \val^i_F\in\reals{r_k\times |\nodes'|}$ denote the $i$-th $k$-headed query, key, and value embeddings of $\feat$ respectively, where $r_k=r/k$, then a multi-headed self-attention Transformer encoder produces features $\feat'$ given by:
\begin{equation}
   \feat' := \concat_{i=1}^k \softmax\left(\frac{\query^i_F {\key^i_F}^{\top}}{\sqrt{r_k}}\right) \val^i_F.
   \label{eq:2}
\end{equation}

\noindent\textbf{\nameTxr:}
We note that in a standard Transformer described in~\eqref{eq:2}, each output feature in $\feat'$ is a mixture embedding of several input features, as decided by the similarity computed within the $\softmax$.
Our key idea in \nameTxr is to use a similarity defined by the spatio-temporal proximity of the graph nodes as characterized by our \name scene graph. For two nodes $\node_1,\node_2 \in \nodes'$, let a similarity kernel $\kernel$ be defined as:
\begin{equation}
    \kernel(\node_1, \node_2 | \sigma_S, \sigma_t) = \exp\!\!\left(-\frac{\enorm{p_{\node_1} - p_{\node_2}}^2}{\sigma_S^2}-\frac{\lnorm{t_{\node_1} - t_{\node_2}}}{\sigma_T}\!\!\right),
    \label{eq:kernel}
\end{equation}
capturing the spatial-temporal proximity between $\node_1$ and $\node_2$ for scales $\sigma_S$ and $\sigma_T$ for spatial and temporal cues, respectively. Then, our \nameTxr is given by:
\begin{equation}
   \ftd' := \concat_{i=1}^k \softmax \Kernel(\nodes', \nodes'|\sigma_S,\sigma_T)\val^i_F,
   \label{eq:4}
\end{equation}
where we use $\Kernel$ to denote the spatio-temporal kernel matrix constructed on $\nodes'$ using~\eqref{eq:kernel} between every pair of nodes. Such a similarity kernel merges features from nodes in the graph that are spatio-temporally nearby -- such as for example, \emph{a person interacting with an object}, or the dynamics of objects in $\graph_d$. Further, the kernel is computed on a union of the graph nodes in $\graph_{s'}$ and $\graph_{d}$, and thus directly captures the interactions between the static and dynamic graphs. 

\noindent\textbf{Hierarchical \nameTxr:}
Note that our \nameTxr in~\eqref{eq:4} captures the spatio-temporal features at a single granularity as defined by $\sigma_S$ and $\sigma_T$. However, we can improve this representation towards a hierarchical abstraction of the scene graph at multiple granularities. Let $\sigma^j_S, \sigma^j_T$, $j=1,\dots, \eta$ be a set of scale, and let $\MLP_j, j=1,\dots, \eta$ be a series of multilayer perceptrons, then combining~\eqref{eq:2} and~\eqref{eq:4}, we define our hierarchical \nameTxr producing features $\fhd$ as:

\begin{align}
    \fhd\!=\!\sum_{j=1}^\eta\MLP_j\concat_{i=1}^k \left(\softmax \Kernel(\nodes'\!, \nodes'|\sigma^j_S,\!\sigma^j_T)\val^i_F\!\right).
    \label{eq:htxr}
\end{align}
\begin{figure}
    \centering
    \includegraphics[width=6cm,trim={11cm 3.5cm 6.5cm 3cm},clip]{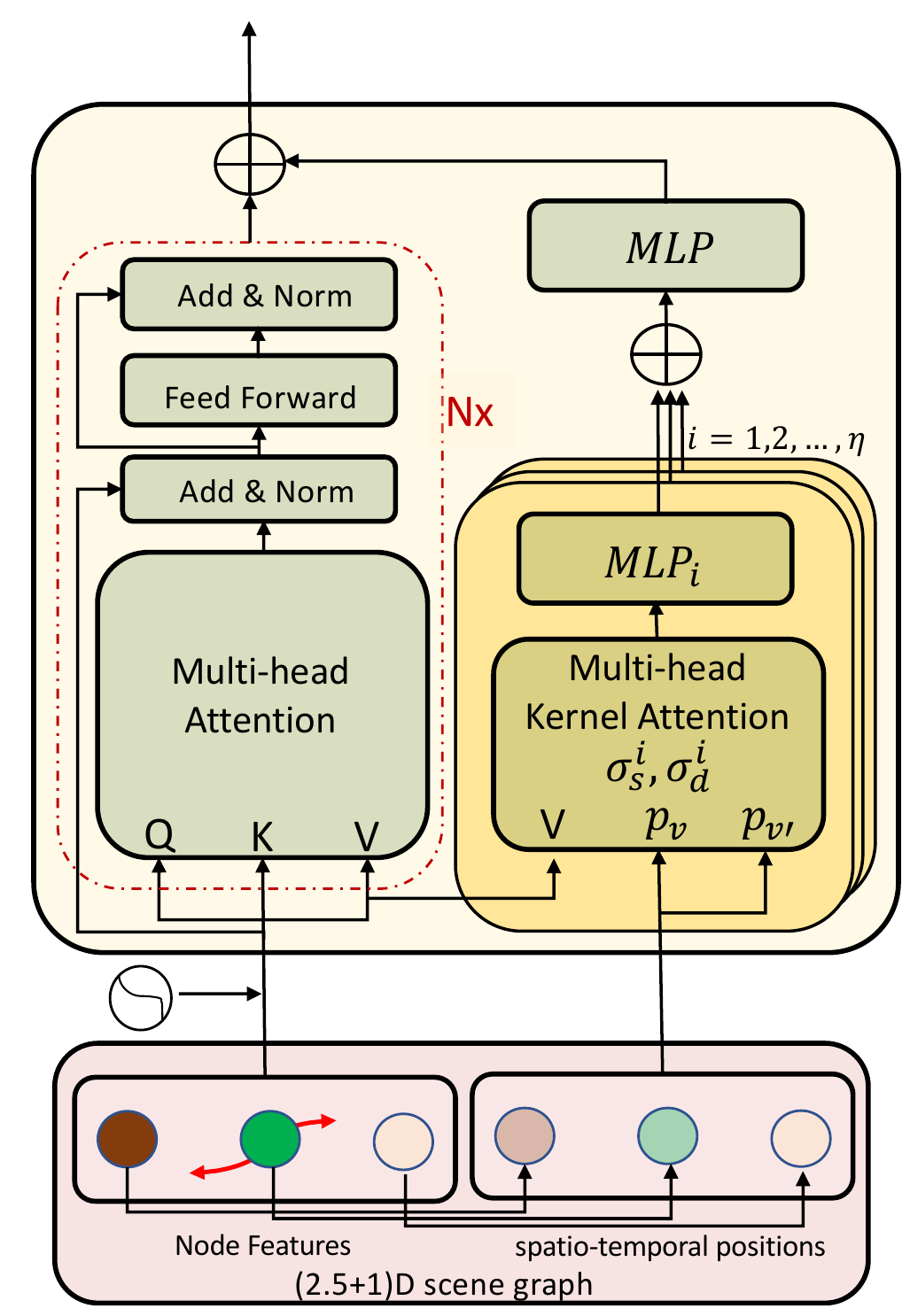}
    \caption{The architecture of the proposed Hierarchical \nameTxr for encoding \name scene graphs. The left module in red (N$\times$) is the standard Transformer. }
    \label{fig:h3.5d_txr}
    \vspace*{-0.5cm}
\end{figure}
In words,~\eqref{eq:htxr} computes spatial-temporal kernels at various bandwidths and merges the respective scene graph node features and embeds them into a hierarchical representation space via the MLPs. In practice, we find that it is useful to combine the kernel similarity in~\eqref{eq:htxr} with the feature similarity in~\eqref{eq:2} and add~\eqref{eq:htxr} with~\eqref{eq:2} (after an $\MLP$) to produce the final graph features. Figure~\ref{fig:h3.5d_txr} shows the architecture of the proposed Transformer.

\noindent\textbf{Question Conditioning:}
For the video QA task, we first use a standard Transformer architecture in~\eqref{eq:2} to produce multi-headed self-attention on the embeddings of the given question. This step precedes attending the encoded questions on $\fhd$ via a multi-headed cross-attention Transformer, followed by average pooling to produce question-conditioned features $\fqd$. In this case, the source to the cross-Transformer is the set of $\fhd$ features, while the target sequence corresponds to the self-attended question embeddings. 

\noindent\textbf{Training Losses:}
To predict an answer $\Ap$ for a given video $\V$ and a question $\Q$, we use the question-conditioned \nameTxr features produced in the previous step and compute its similarities with the set of candidate answers. Specifically, the predicted answer is defined as $\Ap=\softmax({\fqd}{^\top}\lambda(\ansset))$, where $\lambda(\ansset)$ represents embeddings of candidate answers. For training the model, we use cross-entropy loss between $\Ap$ and the ground truth answer $\Agt$. Empirically, we find that rather than computing the cross-entropy loss against one of $\ell$ answers, if we compute the loss against $b\times \ell$ answers produced via concatenating all answers in a batch, that produced better gradients and training. Such a concatenation is usually possible as the text answers for various questions are often different.

\section{Experiments}
In this section, we provide experiments demonstrating the empirical benefits of our proposed representation and inference pipeline. We first review the datasets used in our experiments, following which we describe in detail our setup, before presenting our numerical and qualitative results.

\subsection{Datasets}
We used two recent video QA datasets for evaluating our task, namely NExT-QA~\cite{xiao2021next} and AVSD-QA~\cite{alamri2019audio}. 

\noindent\textbf{NExT-QA Dataset} is a very recent video question answering dataset that goes beyond traditional VQA tasks, and incorporates a significant number of~\emph{why} and \emph{how} questions, that often demand higher level abstractions and semantic reasoning about the videos. The dataset consists of 3,870 training, 570 validation, and 1,000 test videos. The dataset provides 34,132, 4,996, and 8,564 multiple choice questions in the training, validation, and test sets respectively, and the task is to select one of the five candidate answers.  As the test video labels are with-held for online evaluation, we report performances on the validation set in our experiments. We use the code provided by the authors of~\cite{xiao2021next} for our experiments, which we modified to incorporate our Transformer pipeline.

\noindent\textbf{AVSD-QA Dataset} is a variant of the Audio-Visual Scene Aware Dialog dataset~\cite{AVSD@DSTC7}, repurposed for the QA task. The dataset consists of not only QA pairs, but also provides a human generated conversation history, and  captions for each video. In the QA version of this dataset, the task is to use the last question from dialog history about the video to select an answer from one of a hundred candidate answers. The dataset consists of 11,816 video clips and 118,160 QA pairs, of which we follow the standard splits to use 7,985, 1,863, and 1,968 clips for training, validation, and test. We report the performances on the test set. For this dataset, we used an implementation that is shared by the authors of~\cite{geng2021dynamic} and incorporated our modules.

\subsection{Experimental Setup and Training}
\noindent\textbf{Visual Features:} As mentioned earlier, we use public implementations for constructing our scene graphs and \name graphs, these steps being done offline. Specifically, the video frames are sub-sampled at fixed 0.5 fps for constructing the scene graphs using Faster RCNN, and each frame is processed by a MiDAS pre-trained model\footnote{\url{https://pytorch.org/hub/intelisl_midas_v2/}} for computing the RGBD images. The FRCNN and depth images are then combined in a pre-processing stage for pruning the scene graph nodes as described earlier. Out of 1600 object classes in the Visual Genome dataset, we classified 1128 of the classes as \emph{dynamic} and used those for constructing the dynamic scene graph. Next, we used the I3D action recognition model~\cite{carreira2017quo} to extract motion features from the dynamic graph nodes. For this model, we used the videos at their original frame rate, but averaged the spatio-temporal volumes via conditioning on the pruned FRCNN bounding boxes for every dynamic object anchored at the frame corresponding to the frame rate used in the object detection model. This setup produced 2048D features for the static graph nodes and (2048+1024)D features for the dynamic graph nodes. These features are then separately projected into a latent space of 256 for NExT-QA and 128 dimensions for AVSD-QA datasets on which the Transformers operate.

\noindent\textbf{Text Features:} For the NExT-QA dataset, we use the provided BERT features for every question embedding. These are 768D features, which we project into 256D latent space to be combined with our visual features. Each candidate answer is concatenated with the question, and BERT features are computed before matching them with the visual features for selecting the answer. For NExT-QA, we also augment the BERT features with the recent CLIP features~\cite{radford2021learning} that are known to have better vision-language alignment. For AVSD-QA, we used the provided implementation to encode the question and the answers using an LSTM into a 128D feature space. We used the same LSTM to encode the dialog history and the caption features; these features are then combined with the visual features using multi-headed shuffled Transformers as suggested in~\cite{geng2021dynamic}.

\noindent\textbf{Evaluation Protocol:} We used the classification accuracy on NExT-QA, while we use mean retrieval rank on the AVSD-QA dataset; the latter measure ranks the correct answer among the selections made by an algorithm and reports the mean rank over the test set. Thus, a lower mean rank suggests better performance.

\noindent\textbf{Training Details:} We use an Adam optimizer for training both the models. For NExT-QA, we used a learning rate of 5e-5 as suggested in the paper with a batch size of 64 and trained for 50 epochs, while AVSD-QA used a learning rate of 1e-3 and a batch size of 100, and trained for 20 epochs.

\noindent\textbf{Hyperparameters:} There are two key hyperparameters in our model, namely (i) the number of spatial abstraction levels in the hierarchical Transformer, and (ii) the bandwidths for the spatio-temporal kernels. We found that for the NExT-QA dataset, a four layer hierarchy with $\sigma_S\in\set{0.01, 0.1, 1, 10}$ showed the best results, while for
AVSD-QA, we used $\sigma_S\in\set{1,10}$. As for the temporal scale, we divided the frame index $t_\node$ by the maximum number of video frames in the dataset (making the temporal span of the video to be in the unit interval), and used $\sigma_T=\sigma_S$. We found that using a larger number of hierarchical levels did not change the performance for NExT-QA, while it showed slightly inferior performance on AVSD-QA. For the Transformer, we used a 4-headed attention for NExT-QA, and a 2-headed attention for AVSD-QA. 

\subsection{Results}
In this section, we provide numerical results of our approach against state of the art, as well as analyze the contribution of each component in our setup. 

\noindent\textbf{State-of-the-art Comparisons:} In Tables~\ref{tab:sota_nextqa} and~\ref{tab:sota_avsdqa}, we compare the performance of our full \nameTxr pipeline against recent state-of-the-art methods. Notably, on NExT-QA we compare with methods that use spatio-temporal models for VQA such as spatio-temporal reasoning~\cite{jang2019video}, graph alignment~\cite{jiang2020reasoning}, hierarchical relation models~\cite{le2020hierarchical}, against which our proposed model shows a significant $\sim$4\% improvement, clearly showing benefits. On AVSD-QA, as provided in Table~\ref{tab:sota_avsdqa}, we compare against the state of the art STSGR model~\cite{geng2021dynamic}, as well as older multimodal Transformers~\cite{le2019multimodal}, outperforming them in the mean rank of the retrieved answer. We found that when our AVSD-QA model is combined with other text cues (such as dialog history and captions), the mean rank improves to nearly 1.4, suggesting a significant bias between the questions and the text-cues. Thus, we restrict our analysis only to using the visual features.

\begin{table}[]
    \centering
    \sisetup{table-format=2.2,round-mode=places,round-precision=2,table-number-alignment = center,detect-weight=true,detect-inline-weight=math}
    \caption{NExT-QA: Comparisons to the state of the art. Results for the various competitive methods are taken from~\cite{xiao2021next}.}
    \vspace{-.3cm}
    \resizebox{\linewidth}{!}{
    \begin{tabular}{lS}
    \toprule
    Method & {Accuracy (\%)$\uparrow$}\\
    \midrule
       Spatio-Temporal VQA~\cite{jang2019video}  &  47.94\\
       Co-Memory-QA~\cite{gao2018motion} & 48.04\\
       Hier. Relation n/w~\cite{le2020hierarchical}  & 48.20\\
       Multi-modal Attn VQA~\cite{fan2019heterogeneous} & 48.72\\
       graph-alignment VQA~\cite{jiang2020reasoning} &  49.74\\
       \midrule
       \nameTxr (ours) & \bfseries 53.4\\
       \bottomrule
    \end{tabular}
    }
    \label{tab:sota_nextqa}
\end{table}

\begin{table}[]
    \centering
    \sisetup{table-format=1.2,round-mode=places,round-precision=2,table-number-alignment = center,detect-weight=true,detect-inline-weight=math}
    \caption{AVSD-QA: Comparisons to the state of the art. The prior results are taken from~\cite{geng2021dynamic}.}
    \vspace{-.3cm}
    \resizebox{\linewidth}{!}{
    \begin{tabular}{lS}
    \toprule
    Method & {Mean Rank $\downarrow$}\\
    \midrule
       Question Only~\cite{alamri2019audio} & 7.63\\
       Multimodal Transformers~\cite{hori2019end} & 7.23\\
       Question + Video~\cite{alamri2019audio} & 6.86 \\
       MTN~\cite{le2019multimodal} & 6.85\\
       ST Scene Graphs~\cite{geng2021dynamic} &  5.91\\
       \midrule
       \nameTxr (ours) & \bfseries 5.84\\
       \bottomrule
    \end{tabular}
    }
    \label{tab:sota_avsdqa}
\end{table}

\subsection{Ablation Studies}
In Table~\ref{tab:abl_nextqa_avsdqa}, we provide an ablation study on the importance of each component in our setup on both the datasets. Our results show that without the static or the dynamic subgraphs, the performance drops. Without I3D features, the performance drops significantly for both the datasets, underlining the importance of motion features in the graph pipeline. We find that without the hierarchical Transformer, the performance drops from 53.4$\to$52.9 on NExT-QA, and 5.84$\to$5.97 on AVSD-QA. Further, the trick of using augmented answers in the learning process as described in the section on Training Losses seems to help improve the training of the models. We also evaluate the importance of question conditioning, which appears to contribute to the final performance. As our proposed pipeline is sequential in nature, we may also study the performance via removing individual modules from the pipeline. In Table~\ref{tab:nextqa-ablation} rows 1-4, we show the results of this experiment. Our results show that our proposed Transformer module leads to nearly 3\% improvement, and using the pseudo-depth improves by a further 1\%. 

In Table~\ref{tab:hier-ablation}, we ablate on the performance of the hierarchy in the \nameTxr. Specifically, we show results for various bandwidths and their combinations on the NExT-QA dataset. The results suggest that including more bandwidths (hierarchies) leads to better modeling of the video scene and better performance. We use only the \nameTxr for this experiment without using the combination with the standard Transformer to clearly separate out the benefits.

\noindent\textbf{Computational Benefits:} In Table~\ref{tab:nextqa-ablation} last row, we further show ablations on NExT-QA dataset, when the full set of graph nodes are used for inference. As expected in this case, the performance improves mildly, however our experiments show that the time taken for every training iteration in this case slows down 4-fold (from $\sim$1.5 s per iteration to $\sim$6 s on a single RTX6000 GPU). In Table~\ref{tab:abl_compute}, we compare the number of nodes in the static and dynamic graphs, and compare it to the total number of nodes in the unpruned graph for both the datasets. As the results show, our method prunes nearly 54\% of graph nodes on AVSD-QA dataset and 24\% on NExT-QA. We believe the higher pruning ratio for AVSD-QA is perhaps due to the fact that most of its videos do not contain shot-switches and use a stationary camera, which is not the case with NExT-QA.

\begin{table}[]
    \centering
    \sisetup{table-format=2.2,round-mode=places,round-precision=2,table-number-alignment = center,detect-weight=true,detect-inline-weight=math}
    \caption{Ablation study on NExT-QA and AVSD-QA. Below, \emph{Txr} is the standard Transformer and \emph{I3D+FRCNN} is the averaged I3D and FRCNN features per frame (no graph), \emph{V(2+1)D Txr}: without depth.}
    \vspace{-.3cm}
    \setlength{\tabcolsep}{10pt}
    \resizebox{0.9\linewidth}{!}{
    \begin{tabular}{lSS[table-format=2.1]}
    \toprule
     & {NExT-QA} & {AVSD-QA}\\
     \cmidrule(lr){2-2}\cmidrule(lr){3-3}
    Method & {Acc (\%)$\uparrow$} & {mean rank$\downarrow$}\\
    \midrule
       No dynamic graph & 52.49 & 5.97\\
       No static graph & 53.0 & 6.03\\
       No I3D & 52.65 & 6.09\\
       No hier. kernel & 52.9 & 5.97\\
       No ans. augment (AA) & 49.98 & 5.92 \\
       No question condition (QC) & 50.39 & 5.96\\
       \midrule
       Full Model & \bfseries 53.4 & \bfseries 5.84\\
       \bottomrule
    \end{tabular}
    }
    \label{tab:abl_nextqa_avsdqa}
\vspace*{0.1cm}
    \centering
    \sisetup{table-format=2.1,round-mode=places,round-precision=2,table-number-alignment = center,detect-weight=true,detect-inline-weight=math}
    \caption{Ablation study on NExT-QA by removing modules in our pipeline.}
    \setlength{\tabcolsep}{10pt}
    \resizebox{0.9\linewidth}{!}
    {
    \begin{tabular}{llS}
    \toprule
     \# & Ablation & {Accuracy (\%)$\uparrow$} \\
       \midrule
    1 & Txr + I3D + FRCNN + QC & 47.9\\
    2 & (1) + AA & 49.8 \\
    3 & Txr + V(2+1)D Txr + AA + QC & 52.4\\
    4 & Txr + V(2.5+1)D Txr + AA + QC & 53.4\\
    \hline
    5 & (4) using all nodes (no pruning) & \bfseries 53.5\\
     \bottomrule
    \end{tabular}
    }
\label{tab:nextqa-ablation}
\vspace*{0.1cm}
    \centering
    \sisetup{table-format=2.1,round-mode=places,round-precision=2,table-number-alignment = center,detect-weight=true,detect-inline-weight=math}
    \caption{Ablation study on NExT-QA using different spatio-temporal hierarchies defined by the kernel bandwidth $\sigma$.}
    \vspace{-.1cm}
    \setlength{\tabcolsep}{10pt}
    \resizebox{0.9\linewidth}{!}
    {
    \begin{tabular}{llS}
     Hier. levels & bandwidths $\sigma$ & Accuracy \\
    \hline
    1-level & $0.01$ & 52.13 \\
    2-levels & $\set{0.01, 0.1}$ & 52.58\\
    3-levels & $\set{0.01, 0.1, 1.0}$ & 52.97 \\
    4-levels & $\set{0.01,  0.1, 1.0, 10.0}$ & 53.20\\
    5-levels & $\set{0.01, 0.1, 1.0, 10, 20.0}$ & 53.0\\
    \end{tabular}
    }
\label{tab:hier-ablation}
\vspace*{0.1cm}
    \centering
         \sisetup{table-format=3.2,round-mode=places,round-precision=2,table-number-alignment = center,detect-weight=true,detect-inline-weight=math}
    \caption{Computational benefits of the proposed approach. The numbers indicate the average number of graph nodes per video sequence in each dataset.}
    \vspace{-.1cm}
    \setlength{\tabcolsep}{10pt}
    \resizebox{0.9\linewidth}{!}{
    \begin{tabular}{lSS}
    \toprule
     & {AVSD-QA} & {NExT-QA}\\
    \midrule
    Full graph    & 502.43  & 656.30\\
    Static graph  & 97.261   & 68.68\\
    Dynamic graph &136.10 & 430.83\\
    \midrule
    \% node reduction & {$53.6$} & {$23.9$}\\
    \bottomrule
    \end{tabular}
    }
    \label{tab:abl_compute}
\end{table}
\begin{table*}[]
    \centering
    \sisetup{table-format=18,round-mode=places,round-precision=2,table-number-alignment = center,detect-weight=true,detect-inline-weight=math}
    \caption{Comparison of our answer selection on various categories in NExT-QA dataset against other recent methods. The numbers for competing methods are taken from~\cite{xiao2021next}.}
    \vspace{-.3cm}
    \setlength{\tabcolsep}{4pt}
    \resizebox{0.99\linewidth}{!}{
    \begin{tabular}{lcccccccccccS}
    \toprule
    Method &Why (W)  &  How (H) &  Avg. (W+H) & Prev\&Next (P\&N) & Present (P) &  Avg. (P\&N+P) &  Count (C) &  Location (L) &  Other (O)  & Avg. (C+L+O)&  Overall\\
    \midrule
    STVQA, IJCV'19 & 45.37 & 43.05 &44.76 &47.52 &51.73 &49.26 &43.50 &65.42 &53.77 &55.86 &47.94\\
    CoMem, CVPR'18 &  46.15 & 42.61 & 45.22 & 48.16 & 50.38 & 49.07 & 41.81 & 67.12& 51.80 &55.34 &48.04\\
    HCRN, CVPR'20 & 46.99 & 42.90 & 45.91 & 48.16  & 50.83 & 49.26 & 40.68  & 65.42 & 49.84 & 53.67 & 48.20\\
    HME, CVPR'19  & 46.52 & 45.24 & 46.18 & 47.52 & 49.17 & 48.20 & 45.20 & 73.56 & 51.15 & 58.30 & 48.72\\
    HGA, AAAI'20 &  46.99  & 44.22  & 46.26 & 49.53 &  52.49 &  50.74  & 44.07  &  72.54  &  55.41  & 59.33 &  49.74 \\
  \midrule
    Ours & \textbf{52.39} & \textbf{48.36} & \textbf{51.33}  & \textbf{50.91} &  \textbf{54.28} & 52.30  & \textbf{46.02}  & \textbf{77.08}  & \textbf{58.31} & \textbf{62.58} & \textbf{53.4}\\
    \midrule
    \% improvement & \cc{+5.4} & \cc{+3.12} & \cc{+5.07} &\cc{+1.38} & \cc{+1.79} & \cc{+1.56}&\cc{+0.82} & \cc{+3.52} & \cc{+2.91} &\cc{+3.25} & \cc{+3.66}\\
    \bottomrule
    \end{tabular}
    }
    \label{tab:sota_categories}
\end{table*}
\begin{figure*}[h]
    \centering
    \includegraphics[width=14.5cm,trim={0cm 13.5cm 3.5cm 3cm},clip]{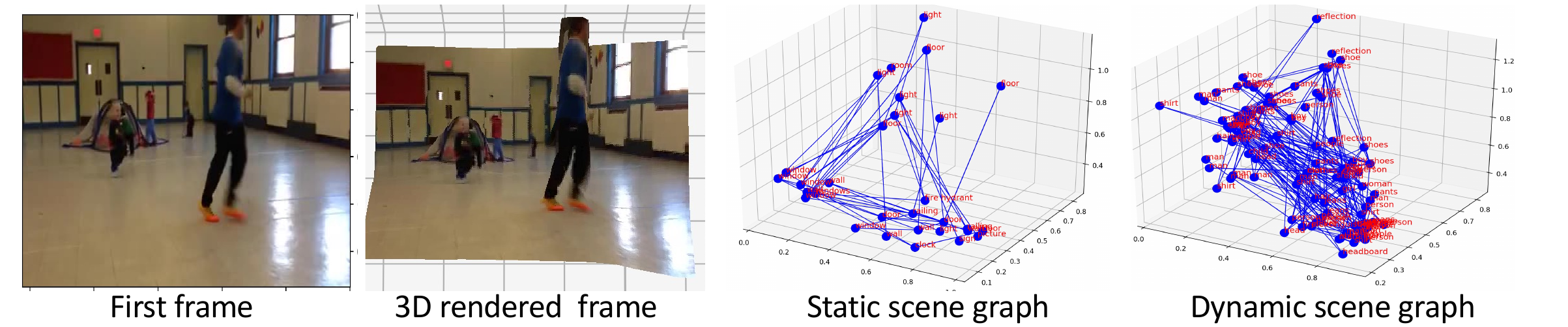}
    \caption{An example illustration of (2.5+1)D scene graphs produced by our method. The figure shows a video frame from the NExT-QA dataset, its pseduo-3D rendering, and the (2.5+1)D static and dynamic graphs computed on all frames of the video.}
    \label{fig:next-qa-more-quals}
    \centering
     \includegraphics[width=14.5cm,trim={0.5cm 9cm 3.5cm 3cm},clip]{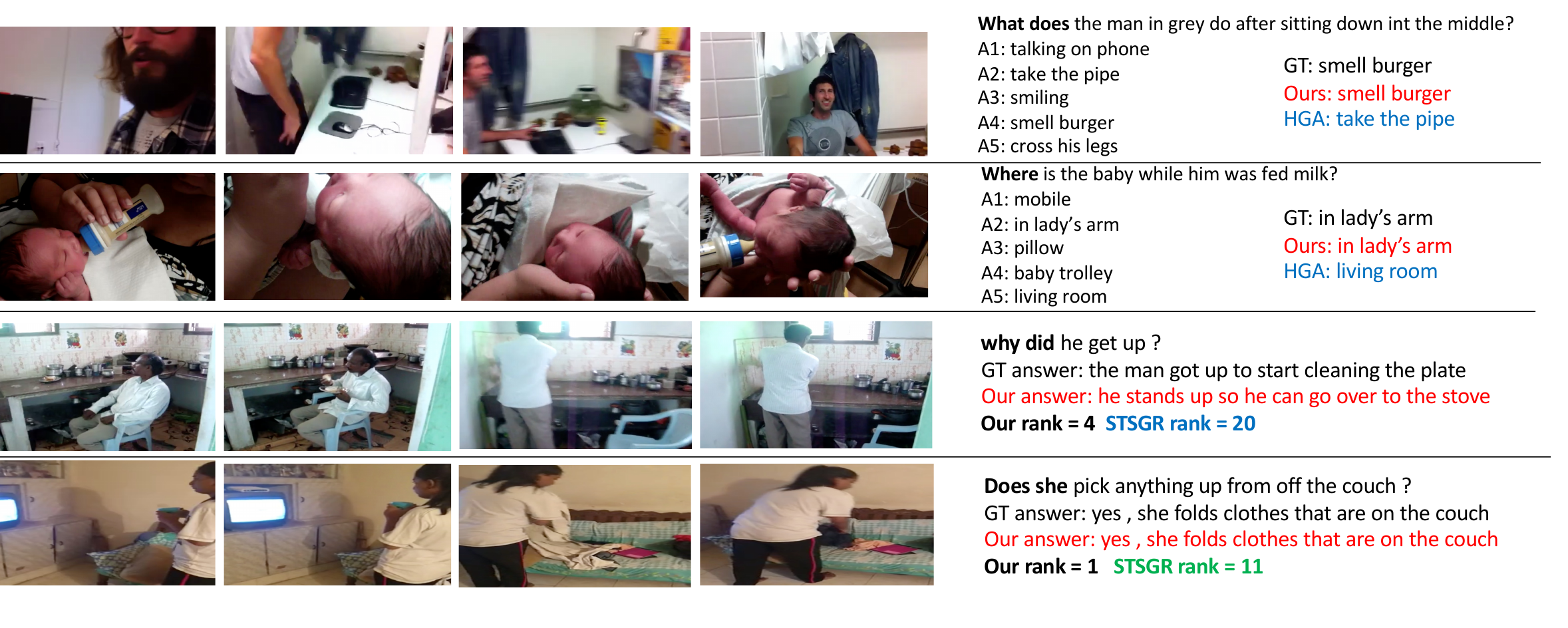}
    \caption{Qualitative responses: First two rows show the results on the NExT-QA dataset and the last two on the AVSD-QA dataset. We compare our results to those produced by HGA~\cite{fan2019heterogeneous} on NExT-QA, and STSGR~\cite{geng2021dynamic} on AVSD-QA datasets. }
    \label{fig:quals}
\end{figure*}
\noindent\textbf{Question-Category Level Performance:} In Table~\ref{tab:sota_categories}, we compare the performance of our \name approach on various question categories in the NExT-QA dataset. Specifically, the dataset categorizes its questions into 7 reasoning classes: (i) why, (ii) how, (iii) previous\&next, (iv) present (v) counting, (vi) spatial location related, and (vii) all other questions. From Table~\ref{tab:sota_categories}, we see that our proposed representation fares well in all the categories against the state of the art. More interestingly, our method works significantly outperforms to the next best scheme HGA~\cite{jiang2020reasoning}, by more than 5\% on \emph{why}-related questions and 3.5\% on \emph{location}-related questions, perhaps due to better spatio-temporal localization of the objects in the scenes as well as the spatio-temporal reasoning.

\begin{figure*}
    \centering
    \includegraphics[width=16cm,trim={0cm 3cm 0cm 3.0cm},clip]{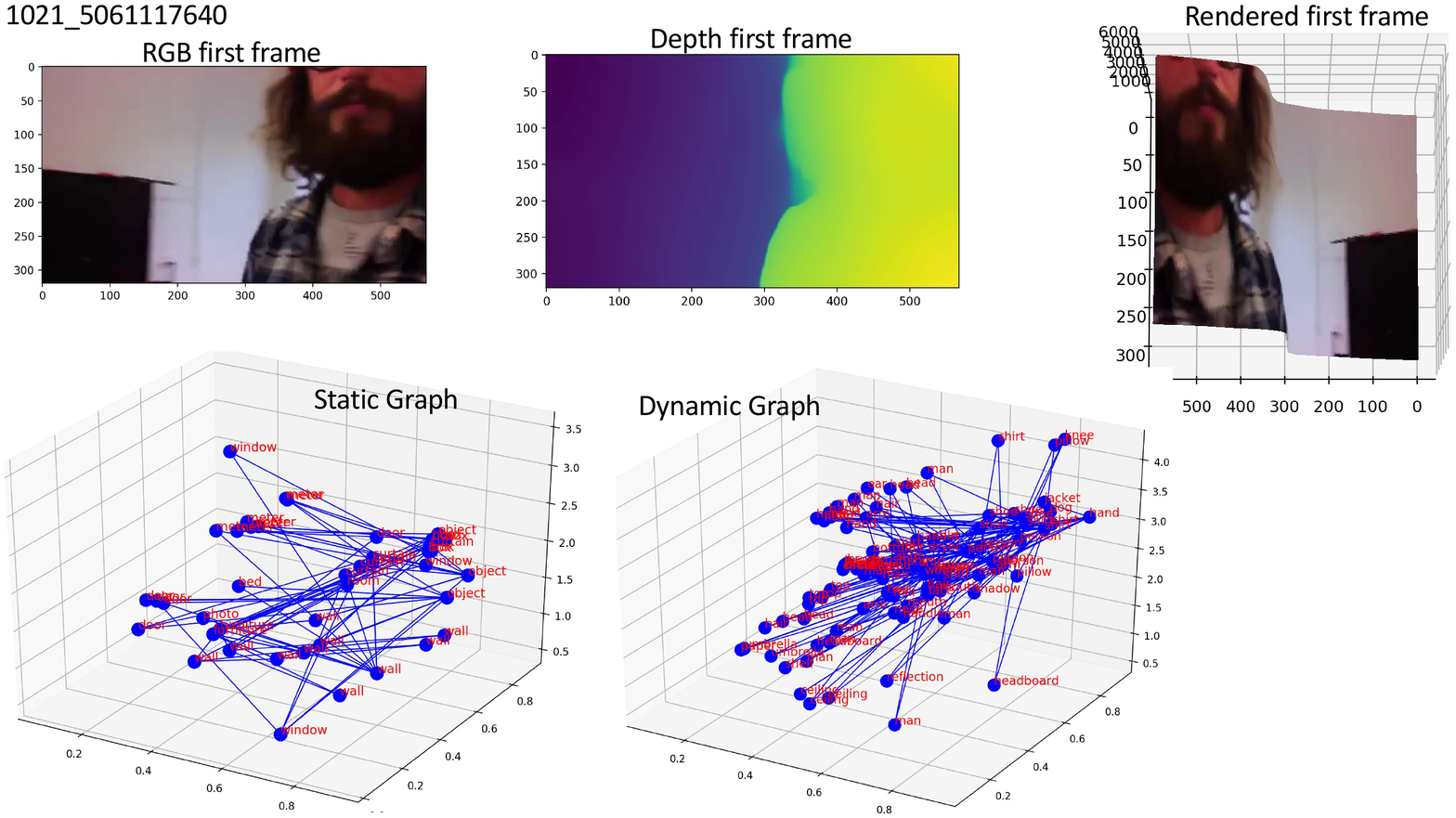}
    \vspace*{10pt}
     \includegraphics[width=16cm,trim={0cm 3cm 0cm 3.0cm},clip]{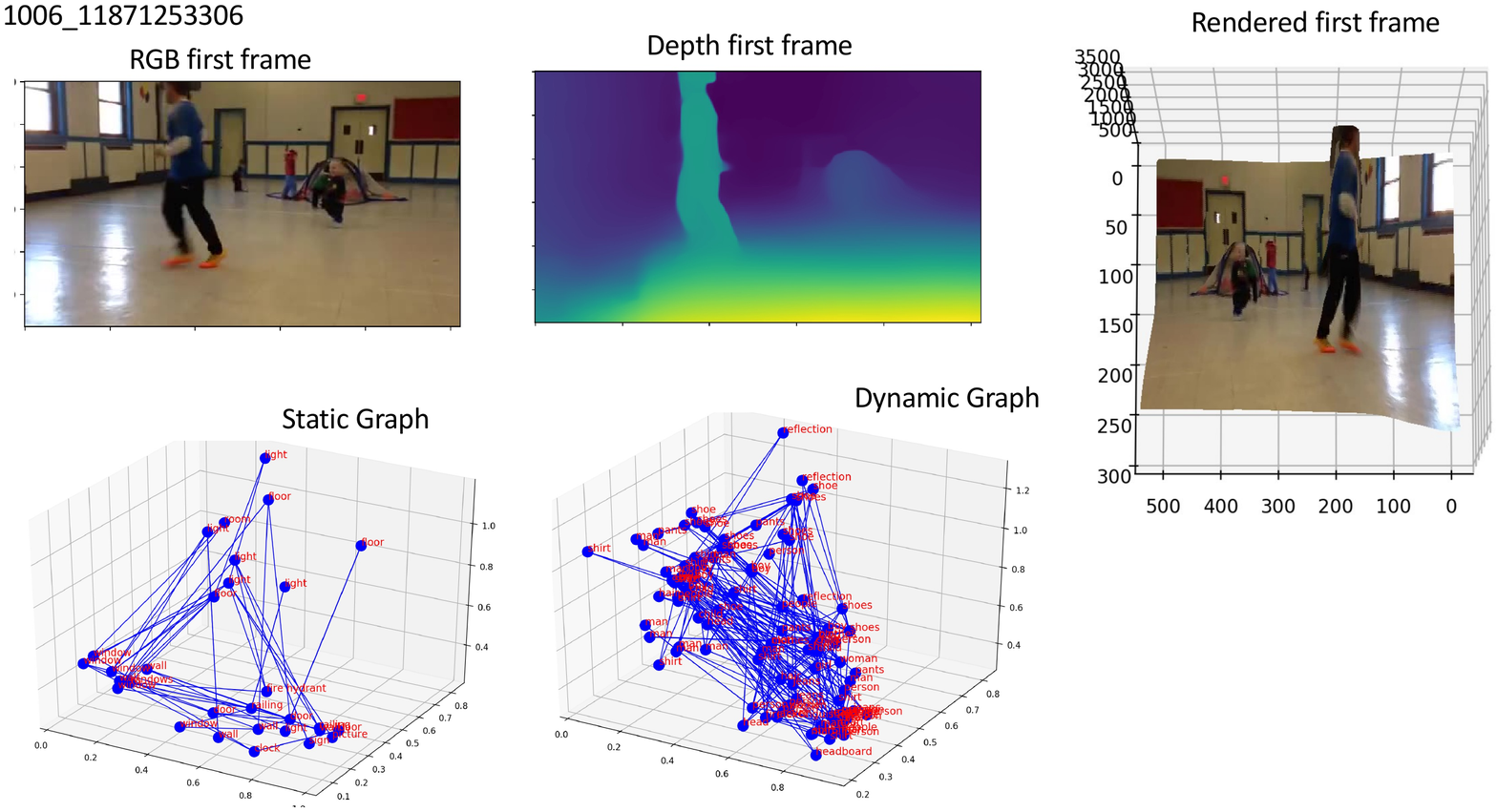}
    \caption{RGB images, depth images produced using~\cite{ranftl2019towards}, the depth rendered RGB images, the static scene graphs, and the dynamic scene graphs for the respective videos. See below for the questions and answers for these videos.}
    \label{fig:quals_1}
\end{figure*}

\begin{figure*}
    \centering
    \includegraphics[width=16cm,trim={0cm 3cm 0cm 3.0cm},clip]{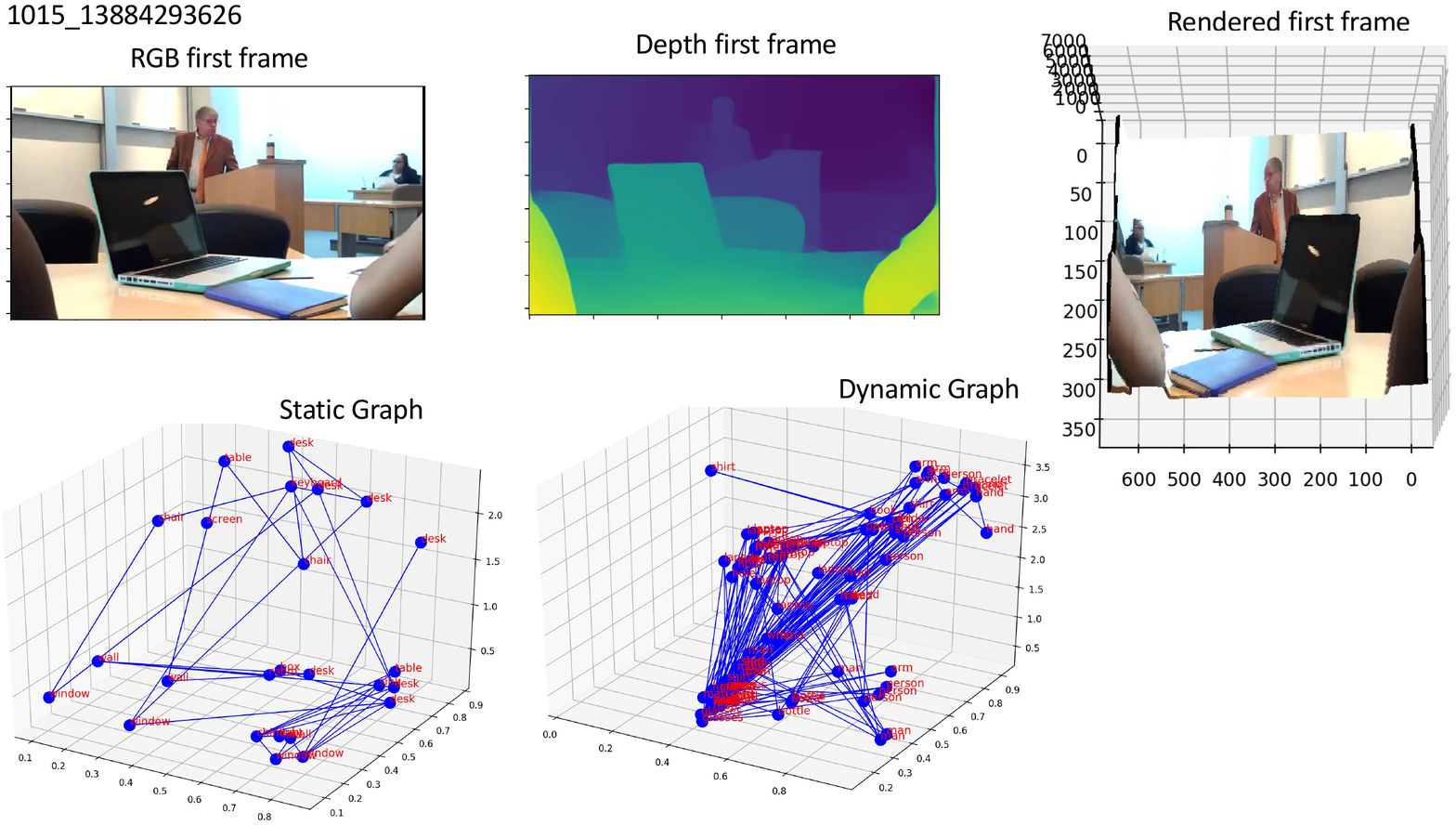}
     \includegraphics[width=16cm,trim={0cm 3cm 0cm 3.0cm},clip]{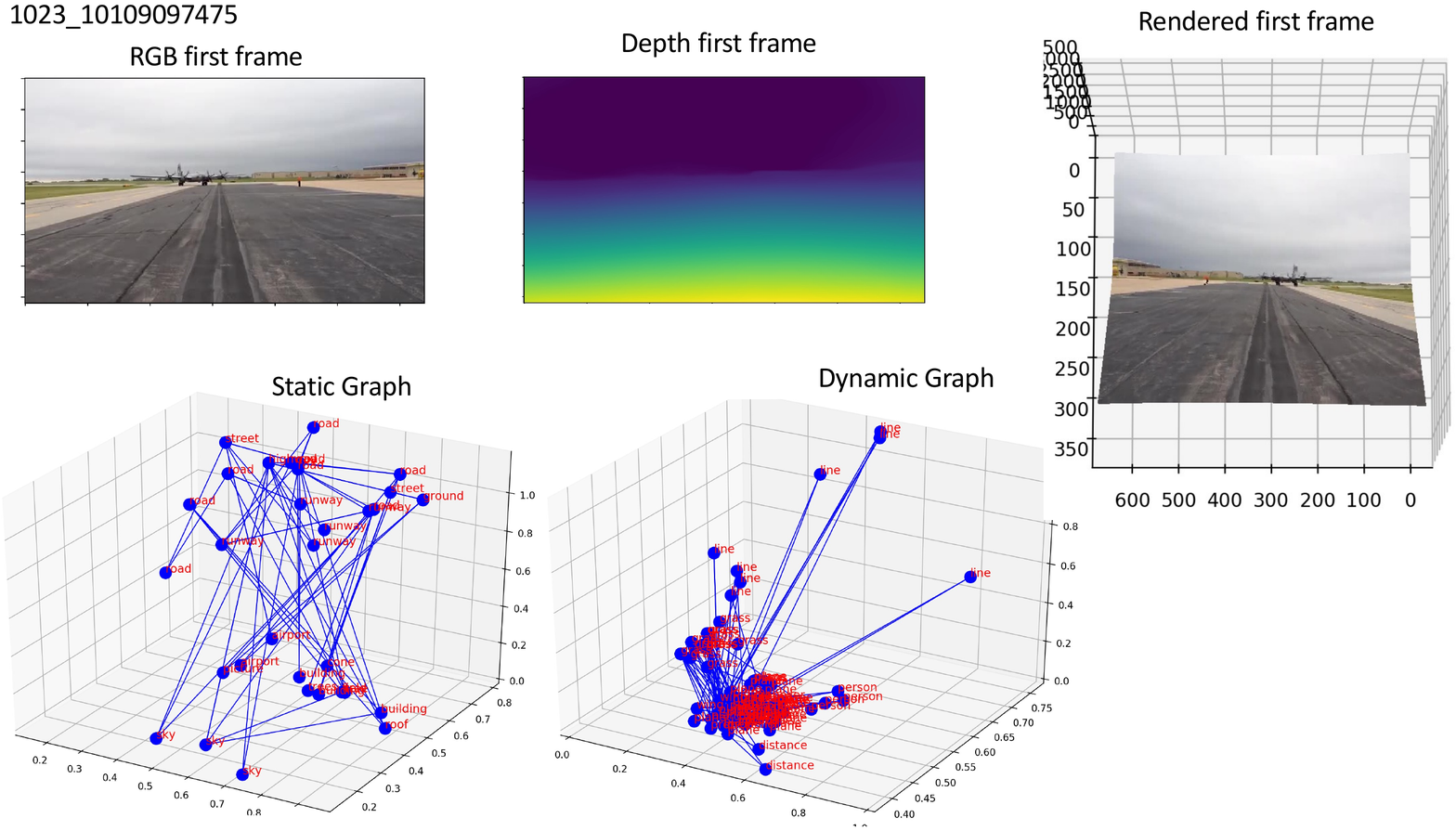}
    \caption{RGB images, depth images produced using~\cite{ranftl2019towards}, the depth rendered RGB images, the static scene graphs, and the dynamic scene graphs for the respective videos. See below for the questions and answers for these videos.}
    \label{fig:quals_2}
\end{figure*}

\begin{figure*}
    \centering
    \includegraphics[width=20cm,trim={0cm 5cm 1.5cm 3cm},clip]{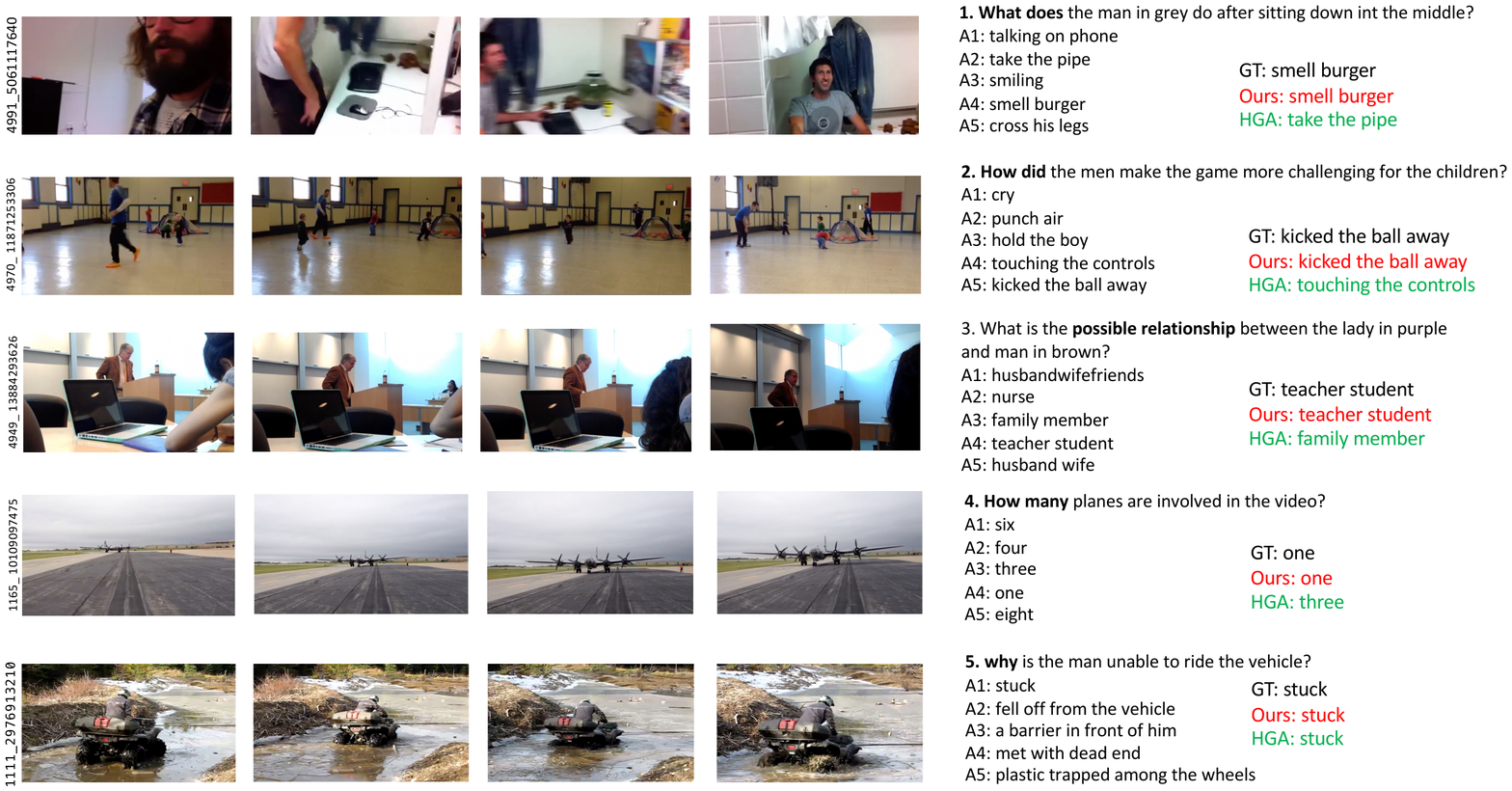}
    \includegraphics[width=20cm,trim={0.5cm 5cm 1.5cm 3cm},clip]{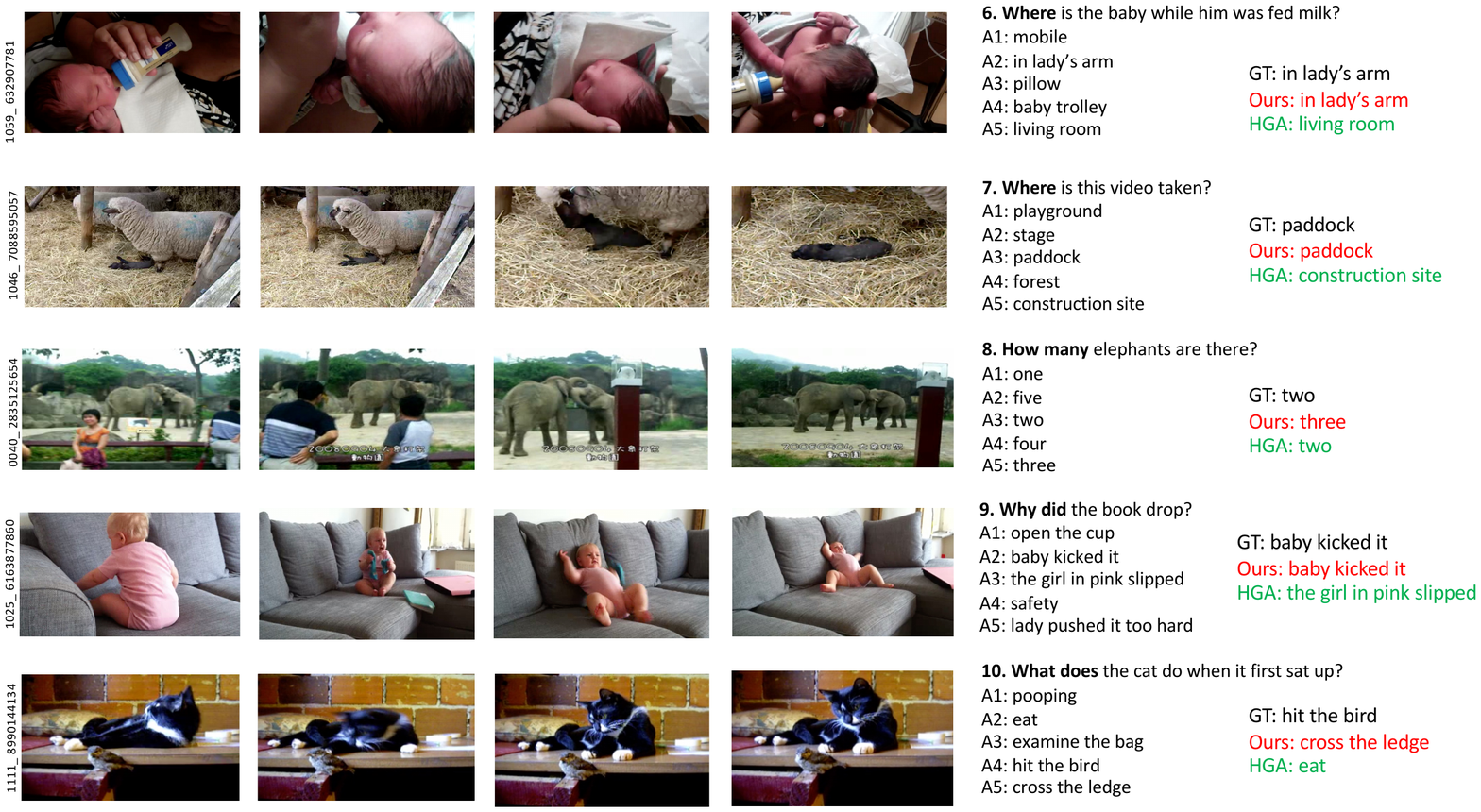}
    \caption{Qualitative responses from the NExT-QA dataset for various types of questions.}
    \label{fig:next-qa-more-quals-x}
\end{figure*}

\begin{figure*}
    \centering
    \includegraphics[width=18cm,trim={1cm 4.5cm 1.5cm 3cm},clip]{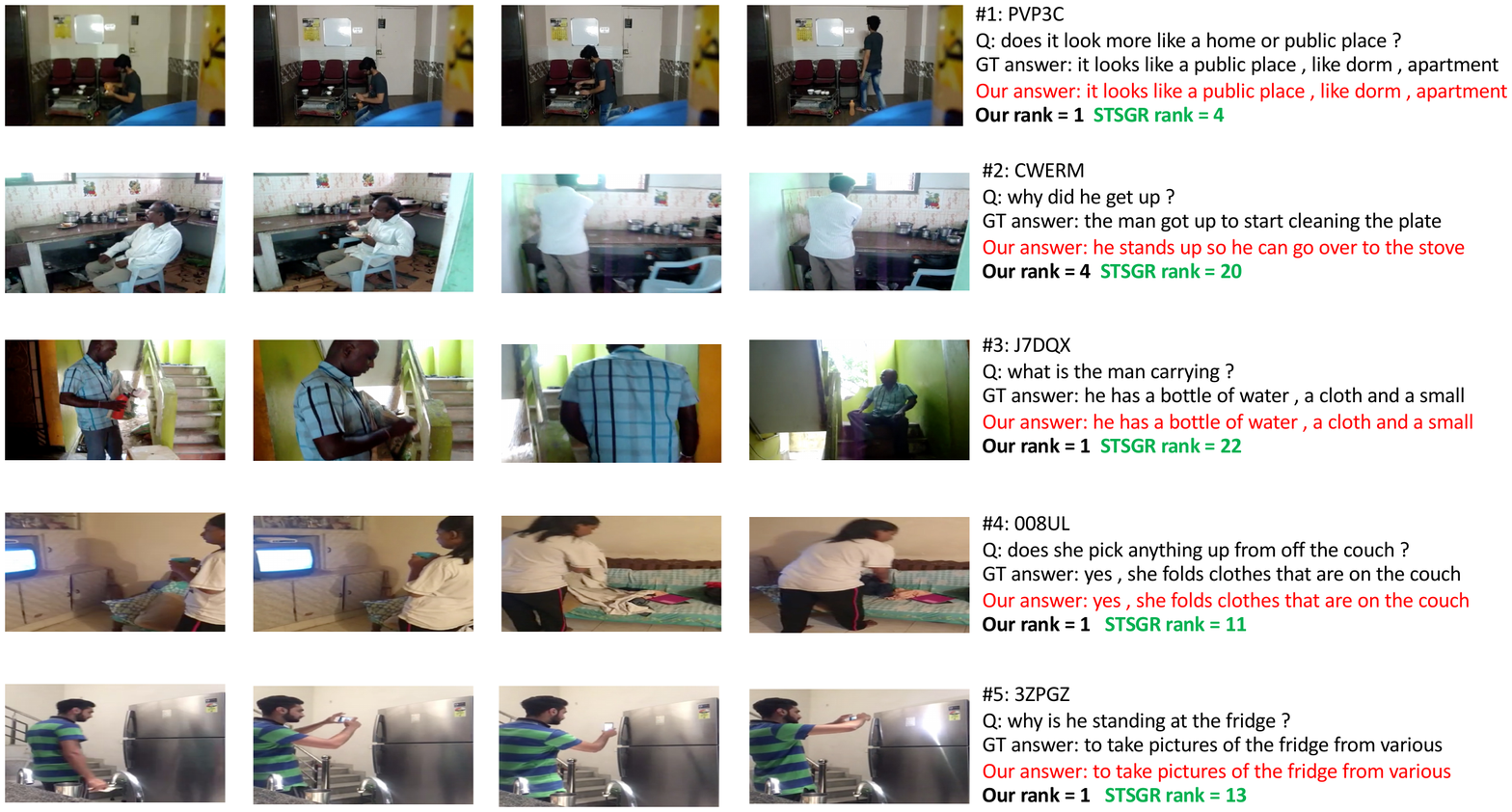}
    \includegraphics[width=18cm,trim={1.5cm 4cm 1.5cm 3cm},clip]{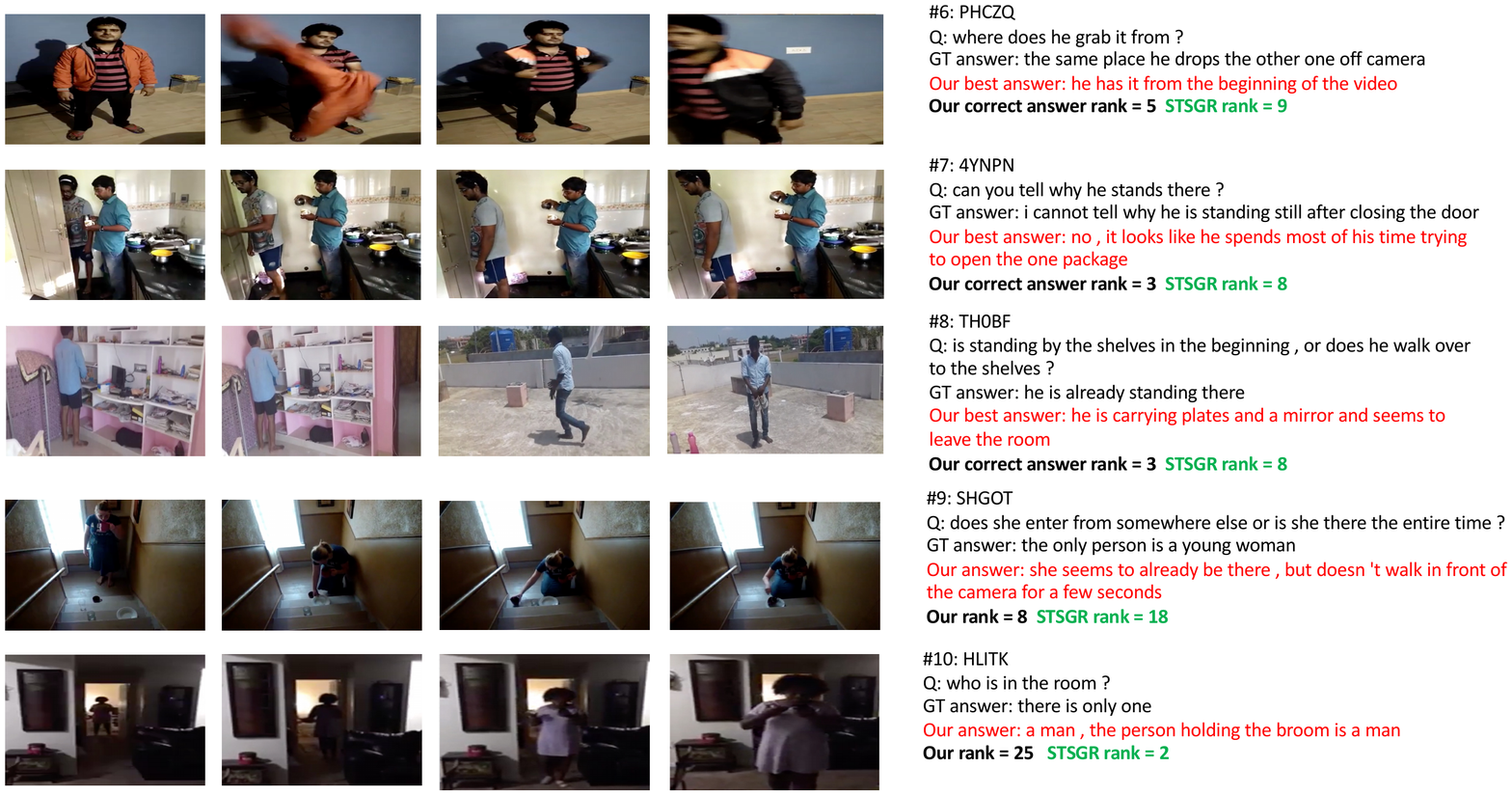}
    \caption{Qualitative responses from the AVSD-QA dataset for various types of questions.}
    \label{fig:avsd-qa-more-quals-x}
\end{figure*}

\noindent\textbf{Qualitative Results:} Figure~\ref{fig:next-qa-more-quals} gives an example static-dynamic scene graph pair on a scene from NExT-QA. In Figure~\ref{fig:quals}, we present qualitative QA results and compare against the responses produced by two recent methods.  See  Figures~\ref{fig:quals_1},~\ref{fig:quals_2},~\ref{fig:next-qa-more-quals-x}, and~\ref{fig:avsd-qa-more-quals-x} for more results.

\section{Conclusions}
In this paper, we presented a novel \name representation for the task of video question answering. We use 2.5D pseudo-depth of scene objects to be disentangled in 3D space, allowing the pruning of redundant detections. Using the 3D setup, we further disentangled the scene into a set of dynamic objects that interact within themselves or with the environment (defined by static nodes); such interactions are characterized in a latent space via spatio-temporal hierarchical transformers, that produce varied abstractions of the scene at different scales. Such abstractions are then combined with text queries in the video QA task to select answers. Our experiments demonstrate state-of-the-art results on two recent and challenging real-world datasets. 

\clearpage
\bibliography{aaai22}

\end{document}